\documentclass[a4paper,fleqn,noorcid]{cas-sc}  
\usepackage[authoryear,longnamesfirst]{natbib}

\usepackage{caption}
\usepackage{cas-common}
\usepackage{graphicx}
\usepackage{diagbox}
\usepackage{mathtools}
\usepackage{amsmath}
\usepackage{bbding}
\usepackage{pifont}
\usepackage{wasysym}
\usepackage{tikz}
\usepackage{calc}
\usepackage{multirow}
\usepackage{rotating}
\usepackage{colortbl}
\usepackage{color}
\usepackage{ragged2e}
\usepackage{float}
\usepackage{bm}
\usepackage{siunitx}
\renewcommand{\printorcid}{}  
\begin{document}
\captionsetup[figure]{labelfont={bf},labelformat={default},name={Fig.}, labelsep=period,}
\captionsetup[table]{labelfont={bf},labelformat={default},labelsep=newline,name={Table}, singlelinecheck=false}
\let\WriteBookmarks\relax
\def\floatpagepagefraction{1}
\def\textpagefraction{.001}
\def\ourmethod{Bidirectional Adaptive Transformers}
\newcommand{\jyx}[1]{\textcolor{black}{#1}}
\shorttitle{Bidirectional Adaptive Transformers for Multimodal Anomaly Detection}    


\title [mode = title]{Bidirectional Adaptive Transformers for Multimodal Anomaly Detection}

\author[1]{Yuxin Jiang}[type=editor, auid=000, bioid=1]
\ead{jiangyuxin@hust.edu.cn}

\author[1]{Yunkang Cao}[type=editor, auid=000, bioid=1]
\ead{caoyunkang@ieee.org}

\author[1]{Weiming Shen}[type=editor, auid=000, bioid=1]
\cormark[1]  
\ead{shenwm@hust.edu.cn}
\cortext[1]{Corresponding author}

\affiliation[1]{organization={School of Mechanical Science and Engineering},
            addressline={Huazhong University of Science and Technology}, 
            city={Wuhan},
            postcode={430000}, 
            state={Hubei},
            country={China}}
            

\begin{abstract}
The effective integration of features derived from RGB images and 3D point clouds presents a substantial challenge in the field of multimodal anomaly detection (MAD). \jyx{Existing MAD methods typically rely on static fusion paradigms built upon frozen pre-trained encoders. While these frameworks facilitate spatial alignment, they lack the mechanism to establish latent semantic correspondence, thereby inducing cross-modal interference that degrades anomaly discriminability.} 
To mitigate these limitations, this study proposes a Bidirectional Adaptive Transformers (BAT) framework, which constructs comprehensive and semantically aligned multimodal representations by leveraging complementary attributes across modalities.
\jyx{The BAT framework introduces a Bidirectional Feature Alignment (BFA) module, which employs a mutual predictive mechanism to bridge the gap between 2D textural patterns and 3D geometric structures. This bidirectional interaction explicitly enhances the textural granularity of point cloud descriptors while augmenting the spatial contextual awareness of RGB features.
To ensure that the latent representations possess sufficient informational density for alignment, BAT incorporates a parameter-efficient adaptation architecture. By utilizing lightweight image and point cloud adapters, the framework facilitates specialized fine-tuning of pre-trained models, mitigating the information loss inherent in frozen encoders.}
Extensive experiments conducted on MVTec 3D-AD and Eyecandies confirm that BAT achieves state-of-the-art anomaly detection performance, significantly outperforming existing MAD methods.

\end{abstract}




\maketitle

\section{Introduction}

\begin{figure}[tp]
\centering
\vspace{2mm}
\includegraphics[width=0.95\columnwidth]{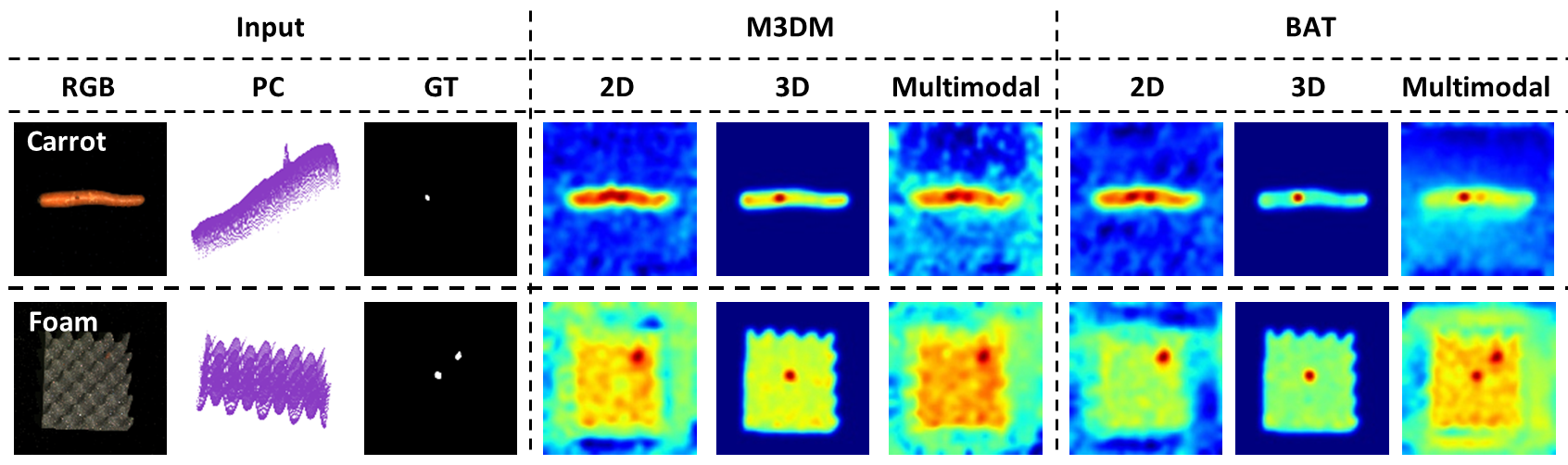}
\caption{\textbf{Anomaly localization examples on MVTec 3D-AD.} The inputs comprise RGB images (RGB), point clouds (PC), and ground truth (GT). Predictions are presented for RGB-based (2D), point cloud-based (3D), and multimodal methods, including M3DM and the proposed BAT. The BAT method demonstrates superior precision in anomaly localization compared to M3DM.}
\vspace{-2mm}
\label{fig:first}
\end{figure}

Anomaly detection, the process of identifying instances that deviate from normative patterns within a dataset, is pivotal across diverse applications, such as industrial defect detection \citep{TDAD, dong2025simple, wang2025minimaxad}, medical diagnostics \citep{guo2023encoder}, and video surveillance \citep{yuan2024surveillance}. 
\jyx{Traditional anomaly detection has been dominated by RGB-based paradigms, which have achieved remarkable success by leveraging deep convolutional and transformer-based architectures. However, these methods are fundamentally constrained by their reliance on 2D textural cues, making them susceptible to failure in scenarios where anomalies exhibit subtle color variations or lack distinctive textural contrast.} 
As depicted in Fig.~\ref{fig:first}, relying solely on RGB inputs with insufficiently discriminative features often lead to the erroneous classification of normal regions as anomalous, thereby compromising detection accuracy.
To overcome these limitations, recent advancements have incorporated 3D point clouds, which provide rich geometric information to complement RGB inputs \citep{yuan2024learning, yuan2023egst}.
\jyx{This integration of textural attributes from RGB images with structural insights from point clouds has established the paradigm of \emph{multimodal anomaly detection} (MAD) \citep{wang2023multimodal, cao2024complementary}.}

\begin{figure}[tp]
\centering
\vspace{2mm}
\includegraphics[width=0.95\columnwidth]{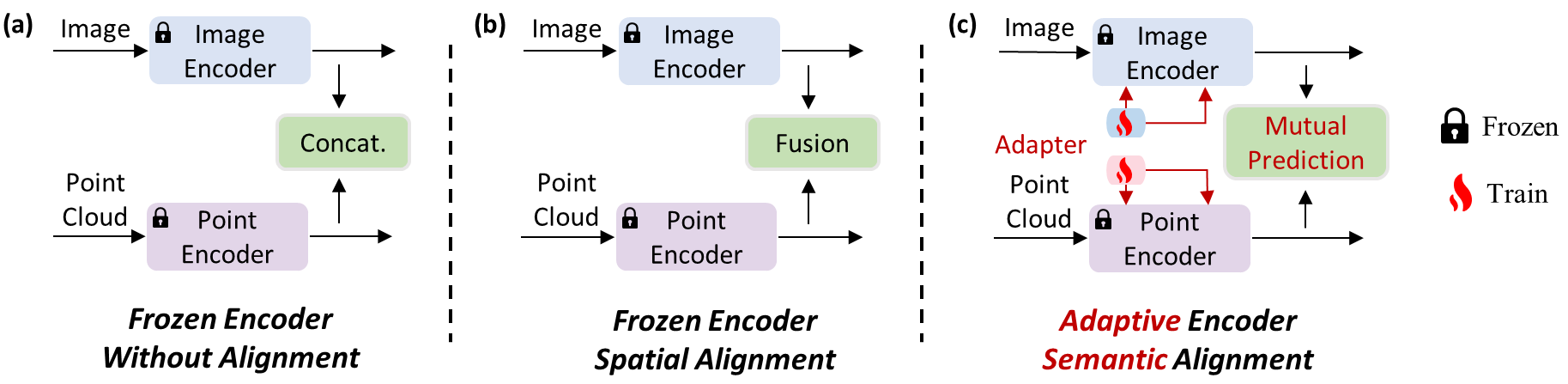}
\caption{\textbf{Comparison of multimodal anomaly detection (MAD) frameworks.} 
(a): Method utilizes straightforward concatenation of images and point clouds with frozen pretrained encoders. 
(b): Method employs spatially aligned fusion with frozen pretrained encoders. 
(c): The proposed BAT framework introduces trainable adapters for both image and point cloud data and leverages mutual prediction to generate semantically aligned multimodal features.}
\vspace{-2mm}
\label{fig:second}
\end{figure}

The effectiveness of MAD relies on the creation of robust, comprehensive feature representations that enable precise anomaly localization. As illustrated in Fig.~\ref{fig:second}(a), early MAD approaches \citep{costanzino2024multimodal, shapenet} typically employed simplistic concatenation of RGB and point cloud features to construct multimodal features. For instance, the Asymmetric Student-Teacher Network (AST) \citep{AST} appends point cloud depth data to RGB channels within a 2D image architecture, which neglects the intricate spatial relationships inherent in 3D data.  \jyx{Similarly, BTF \citep{horwitz2023back} integrates handcrafted 3D descriptors and RGB features via direct concatenation, without explicit feature alignment.} Due to distributional disparities between modalities, such simplistic concatenation strategies fail to fully harness the complementary potential of the modalities, resulting in limited cross-modal synergy.

A significant advancement in MAD was achieved with M3DM ~\citep{wang2023multimodal}, which employs specialized feature fusion modules to enhance cross-modal interactions, as illustrated in Fig.~\ref{fig:second}(b). M3DM utilizes self-supervised pre-trained models, specifically ImageNet \citep{ImageNet} for RGB data and ShapeNet \citep{shapenet} for point clouds, to extract deep features. These features are integrated through an unsupervised fusion mechanism based on spatial correspondence, thereby enhancing representational power. However, its reliance on the straightforward spatial alignment overlooks the critical need for semantic correspondence, limiting its ability to exploit complementary multimodal information. This shortfall is particularly evident in complex scenarios, such as foam samples with concurrent color and shape anomalies or carrots with subtle textural variations, where RGB-driven misjudgments often overshadow essential geometric cues, resulting in cross-modal interference that manifests as either over-detection or under-detection. Furthermore, insights from other anomaly detection frameworks \citep{recontrast} highlight that fixed parameters in pre-trained networks restrict adaptability, resulting in significant information loss during feature extraction, which may compromise robust multimodal semantic alignment.

\jyx{To address these challenges, we propose Bidirectional Adaptive Transformers (BAT) (Fig. \ref{fig:second}c), a framework designed to unify disparate modal representations into a cohesive semantic space. BAT incorporates a Bidirectional Feature Alignment (BFA) module that employs a mutual feature prediction mechanism between RGB and point cloud modalities. This module enhances the textural granularity of point cloud features and the spatial contextual awareness of RGB features. To ensure that the extracted features retain sufficient semantic density for subsequent cross-modal integration, BAT employs lightweight image and point adapters. Rather than relying on frozen, generic representations that may suffer from domain-specific information loss, these adapters fine-tune the backbone to capture modality-specific nuances. This optimization ensures that the inputs to the Bidirectional Feature Alignment (BFA) module are information-rich, providing a robust foundation for establishing precise semantic correspondence between the RGB and geometric domains.}

To validate the multimodal anomaly detection capabilities of BAT, this study conducts extensive experiments on two widely recognized industrial anomaly detection benchmarks: the MVTec 3D-AD and Eyecandies. BAT demonstrates superior performance compared to previous state-of-the-art (SOTA) approaches, achieving an image-level AUROC of 97.0$\%$ on the MVTec 3D dataset.
Furthermore, as illustrated in Figure~\ref{fig:first}, BAT effectively mitigates feature interference resulting from spatial misalignment, exhibiting significantly enhanced localization accuracy compared to M3DM. These findings underscore the robustness of BAT’s semantic alignment strategy in achieving precise anomaly localization.

The contributions of this paper are as follows:

\begin{itemize}
    \item This study introduces the Bidirectional Adaptive Transformers (BAT) framework, a novel method that leverages complementary information from RGB images and 3D point clouds to construct a unified semantic-aligned representation space for enhanced multimodal anomaly detection.
    \item The proposed Bidirectional Feature Alignment (BFA) module leverages mutual prediction mechanisms to enhance feature complementarity, simultaneously refining the textural granularity of point cloud features and augmenting the spatial contextual awareness of RGB features. This strategy ensures semantic alignment, thereby effectively mitigating cross-modal interference.
    \item This study develops lightweight image and point cloud adapters that facilitate efficient fine-tuning of pre-trained models, preserving rich semantic content throughout adaptation to optimize feature quality for subsequent feature alignment in the BFA module.
\end{itemize}

The remainder of this paper is structured as follows: Section~\ref{sec:related_work} offers a thorough exploration of current literature in the field. Section~\ref{sec:method} provides a detailed explanation of the proposed BAT method. Section~\ref{sec:experiment} presents and analyzes evaluation results derived from rigorous assessments on two publicly available datasets. Finally, Section~\ref{sec:conclusion} concludes the paper and discusses potential directions for future research.
\section{Related work}\label{sec:related_work}

\subsection {RGB Anomaly Detection}

\jyx{RGB-based anomaly detection (AD) predominantly operates under an unsupervised learning paradigm \citep{CDO}, where the primary objective is to characterize the underlying manifold of the "normal" data distribution \citep{tao2022deep}. Implicit feature methods leverage generative or reconstruction-based architectures, such as auto-encoders (AEs) \citep{gong2019memorizing}, to map normal images into latent embeddings and subsequently reconstruct the input with high fidelity \citep{ye2020attribute,song2023two,liu2024surface}. As these reconstruction models are trained solely on normal images, they are anticipated to show significant regression errors when presented with abnormal images \citep{liu2023two}.
Explicit feature methods typically rely on models pretrained on large-scale natural image datasets such as ImageNet \citep{krizhevsky2017imagenet} to obtain normal features. These methods quantify normality through various density estimation and statistical techniques, including normalizing flows\citep{gudovskiy2022cflow}, multivariate Gaussian distribution \citep{defard2021padim}, and memory banks \citep{10697302}. These statistical references facilitate the localization of anomalies by measuring the Mahalanobis distance or feature discrepancies \citep{lyu2024reb, liang2023omni}. Furthermore, knowledge distillation (KD) paradigms \citep{zhang2023destseg,deng2022anomaly} train a student network to emulate a pre-trained teacher’s behavior on normal samples. By restricting the student’s capacity to generalize to unseen anomalous patterns, the resulting feature divergence between the teacher and student networks provides a robust signal for anomaly localization \citep{jiang2023masked,cao2022informative, 10970077}.
Fundamentally, these approaches aim to construct a highly discriminative latent space that maximizes the margin between normative and anomalous patterns, thereby establishing a rigorous decision boundary for robust detection.}

\subsection {Multimodal Anomaly Detection}

In real-world scenarios, anomalies often manifest as deviations in both
texture and structural characteristics compared to typical patterns. The
integration of complementary RGB and point cloud modality data facilitates a more
holistic representation of multimodal features, thereby enhancing the understanding of such anomalies. The advent of the first real-world 3D anomaly detection dataset, MVTec 3D-AD \citep{bergmann2021mvtec}, has significantly catalyzed
the interest in multimodal anomaly detection methodologies.
For instance, the Asymmetric Student-Teacher (AST) framework \citep{AST}
employs a normalizing flow as a teacher model, incorporating depth information from point clouds by concatenating it with RGB channels within
an image-based network architecture. However, this approach fails to fully exploit
the complex spatial structures inherent in point cloud data, thereby
limiting its capacity to capture the intricate geometric relationships critical for anomaly detection.
Moreover, BTF \citep{horwitz2023back} explores the effectiveness of traditional handcrafted point cloud data descriptors and underscores the advantages of combining both point cloud and RGB modalities. However, its dependence on manual feature extraction methods and simplistic, unaligned feature concatenation restricts its detection performance.
M3DM \citep{wang2023multimodal} improved over BTF by employing rich and distinctive RGB and point cloud features extracted by frozen transformer-based foundation models, and also proposes an unsupervised spatially aligned feature fusion module to fuse RGB and point cloud features. Despite these advancements, methods that rely on rudimentary concatenation or spatial alignment often fail to establish semantic correspondence, leading to cross-modal interference that undermines detection accuracy.
\jyx{Besides, several approaches adopt cross-modal mapping for anomaly detection. For instance, CFM ~\citep{costanzino2024multimodal} utilizes mapping errors between modalities directly as anomaly scores, operating under the assumption of strict mutual predictability that all normal features in one modality can be faithfully reconstructed from the other.  Meanwhile, methods such as UCF ~\citep{UCF} use cross-modal mapping to generate supplementary embeddings that enrich multimodal representations, combining these with uni-modal signals for detection.
These feature mapping strategies assume mutual predictability between modalities, suggesting that the source modality encodes latent information relevant to the semantic content of the target modality, thereby demonstrating partial semantic overlap and inherent complementarity. However, the unique characteristics of each modality, which cannot be fully reconstructed from the other, may lead to incomplete or inaccurate feature mappings. For instance, fine-grained textural anomalies in RGB data may lack corresponding geometric features in point cloud data, resulting in mapping errors that obscure subtle anomalies.
Moreover, reliance on frozen pretrained networks imposes constraints on feature enhancement, leading to significant information loss during initial feature extraction and hindering effective integration across modalities. 
To address these limitations, the proposed BAT method employs bidirectional alignment to actively refine the feature extractors through trainable adapters. Through mutual prediction between modalities, RGB features acquire spatial awareness from point clouds while point cloud features gain textural granularity from RGB, fundamentally improving representation quality.}
\section{Method}\label{sec:method}

\subsection{Overview}
\begin{figure*}[h]
\centering
\includegraphics[width=0.95\columnwidth]{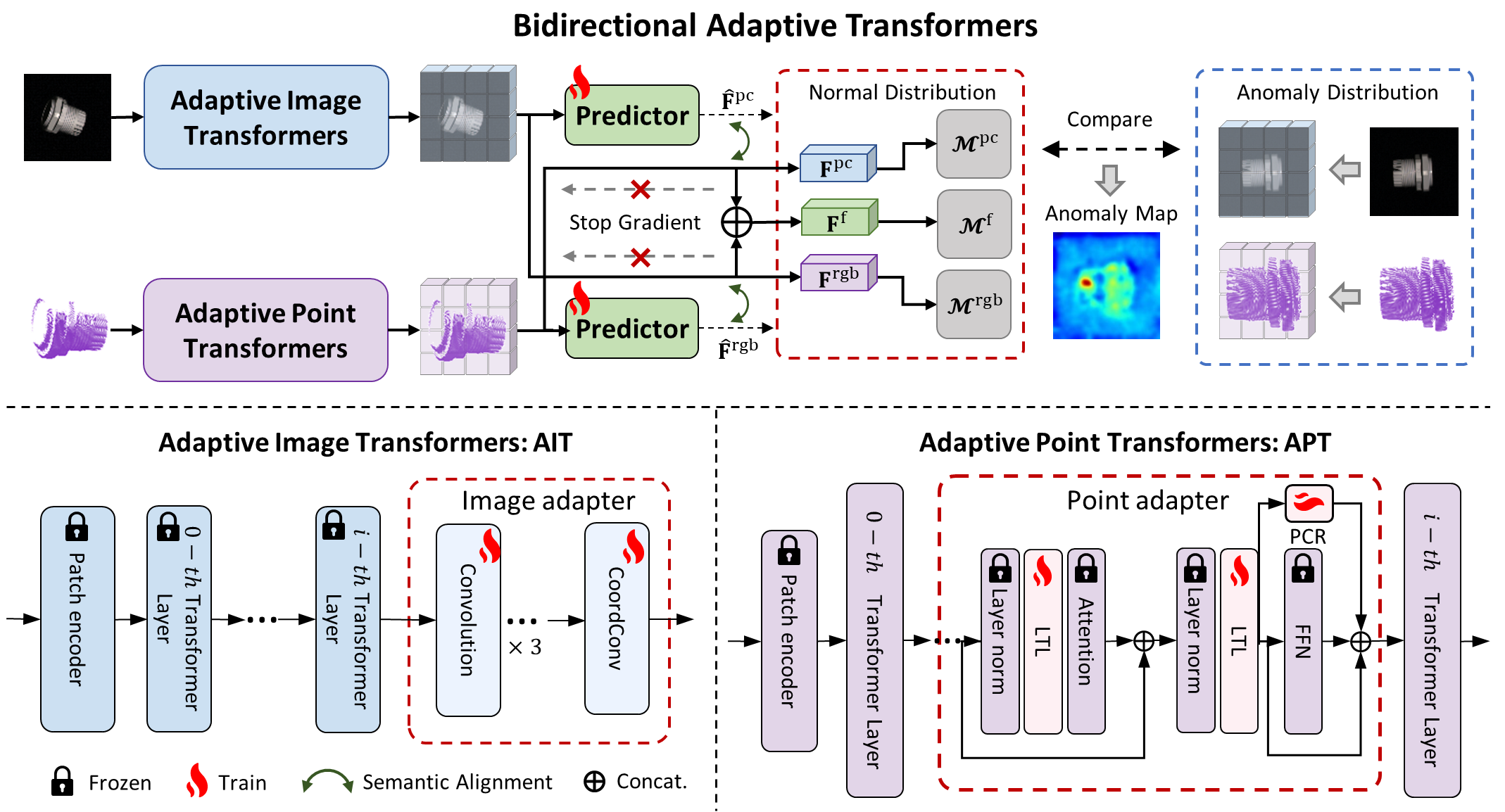}

\caption{
\textbf{Overall Architecture of the BAT Framework.} The BAT framework integrates 2D and 3D processing pipelines through adaptive image and point transformers, generating modality-specific and fused features via bidirectional feature alignment. During testing, these features are stored in memory banks to enable precise multimodal anomaly detection.
}
\label{fig:three}
\end{figure*}

Despite the advancements in MAD methods, obtaining comprehensive multimodal features in industrial contexts presents two primary challenges: (1) the need to promote complementary information learning rather than simple spatial alignment, and (2) the restricted adaptability of frozen pre-trained networks, which leads to significant information loss during initial feature extraction. To address these challenges, this study proposes the Bidirectional Adaptive Transformer (BAT) framework, as illustrated in Figure~\ref{fig:three}. The BAT framework incorporates trainable image and point adapters integrated into pre-trained feature extractors $\phi^{\mathrm{rgb}}$ and $\phi^{\mathrm{pc}}$, yielding adaptive image features $\mathbf{F}^{\mathrm{rgb}}$ and point cloud features $\mathbf{F}^{\mathrm{pc}}$. Additionally, dual predictors are incorporated to facilitate feature prediction between modalities, fostering semantic alignment and generating fused features $\mathbf{F}^{\mathrm{f}}$.
These fused features, along with individual modality features, are stored in memory banks $\mathcal{M}^{\mathrm{rgb}}$, $\mathcal{M}^{\mathrm{pc}}$, and $\mathcal{M}^{\mathrm{f}}$. During the testing phase, the nearest distances between the testing features and the normal feature stored in these memory banks are calculated for anomaly detection.

\subsection{Adaptive Transformers}
To effectively extract features from RGB images and point clouds, this study employs pre-trained transformer models, specifically an image transformer $\phi^{\mathrm{rgb}}$ and a point transformer $\phi^{\mathrm{pc}}$, as backbones. However, the fixed parameters of these pre-trained models often limit their adaptability to downstream tasks, leading to suboptimal feature representations due to the critical information loss during feature extraction. 
To overcome this, this study introduces image and point adapters that enhance fine-tuning capabilities, enabling the extraction of robust multimodal features tailored for MAD tasks.
These adapters integrate with the pre-trained models to form the Adaptive Image and Point Transformer framework, as detailed below.

\noindent\textbf{Adaptive Image Transformer.} 
The Adaptive Image Transformer (AIT) enhances the adaptability of the pre-trained image transformer $\phi^{\mathrm{rgb}}$ while preserving its foundational knowledge. Direct optimization of $\phi^{\mathrm{rgb}}$ risks catastrophic forgetting, potentially degrading the valuable pre-trained knowledge \citep{zhou2024dynamic}. To address this, the AIT incorporates an image adapter comprising three $1 \times 1$ convolutional layers for semantic feature extraction and a CoordConv layer \citep{liu2018intriguing} to enhance spatial awareness. Given an input RGB image $\mathbf{I}$, the AIT generates a feature map $\mathbf{F}^{\mathrm{rgb}}$ tailored to MAD tasks.

\noindent\textbf{Adaptive Point Transformer.} 
To augment the feature extraction capabilities of the pre-trained point transformer $\phi^{\mathrm{pc}}$, we propose a point adapter with a dual-strategy fine-tuning mechanism that combines a Linear Transformation Layer (LTL) and a Point Cloud Refinement (PCR) module. The LTL enables task-agnostic feature modulation, effectively aligning the general feature distribution of the pre-trained model with the downstream task. Complementarily, the PCR utilizes a lightweight nonlinear architecture to dynamically refine deep features, facilitating the capture of intricate details critical for superior MAD performance.

Given an input point cloud $\mathbf{P}$, we partition it into \(N\) non-overlapping patches, transforming it into a 1D sequence $\mathbf{P}_p \in \mathbb{R}^{(N+1) \times d}$, where $d$ is the embedding dimension. This sequence is subsequently processed across \(L\) transformer blocks. Within each block, the LTL executes a lightweight linear transformation to promote efficient fine-tuning \citep{lian2022scaling}. The output $\mathbf{F}_l^{\mathrm{pc}\prime\prime}$ at the $l$-th layer is derived as:
\begin{alignat}{1}
& \mathbf{F}_l^{\mathrm{pc}\prime} = \text{Attention}(\gamma \cdot \text{LN}(\mathbf{P}_p) + \beta) + \mathbf{P}_p,  \\
& \mathbf{F}_l^{\mathrm{pc}\prime\prime} = \gamma \cdot \text{LN}(\mathbf{F}_l^{\mathrm{pc}\prime}) + \beta,
\end{alignat}
where $\text{LN}(\cdot)$ denotes Layer Normalization \citep{ba2016layer}, $\text{Attention}(\cdot)$ denotes self-attention mechanism, and $\gamma, \beta \in \mathbb{R}^{1 \times d}$ are learnable parameters. These parameters introduce minimal computational overhead while significantly enhancing performance.

\begin{figure}
\centering
\includegraphics[width=5cm]{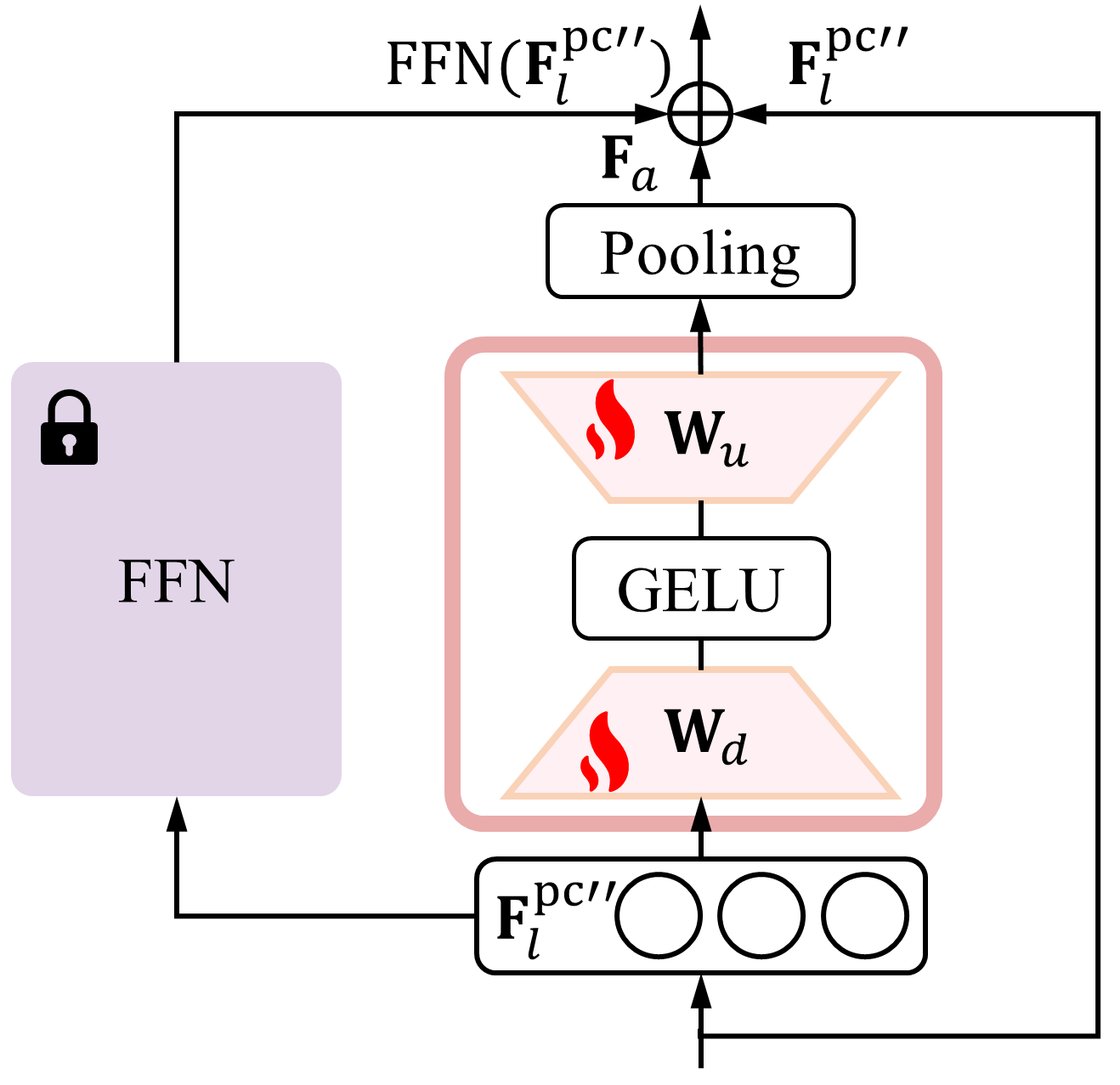}
\caption{\textbf{Architecture of the PCR module.}}
\label{fig:four}
\end{figure}

While LTL provides efficient linear adaptation, its capacity to model complex semantic relationships in point clouds is limited. To address this, we introduce PCR, designed to apply nonlinear transformations for extracting intricate semantic patterns, such as local structural anomalies~\citep{chen2022adaptformer}.
As illustrated in Figure~\ref{fig:four}, the PCR employs a dual-branch design: The left branch preserves the original Feedforward Network (FFN) \citep{dosovitskiy2020image} structure for compatibility with the pre-trained model, while the right branch introduces a bottleneck structure for parameter-efficient fine-tuning.
Specifically, the PCR applies a sequence of projection and pooling operations, with a GELU activation function \citep{GELU} inserted between projection layers to introduce nonlinearity, yielding adapted features:
\begin{equation}
\mathbf{F}_a = \text{GELU}(\mathbf{F}_l^{\mathrm{pc}\prime\prime} \mathbf{W}_d^\top) \mathbf{W}_u^\top \mathbf{W}_p^\top,
\end{equation}
where $\mathbf{W}_d$, $\mathbf{W}_u$, and $\mathbf{W}_p$ represent the down-projection, up-projection, and pooling weights, respectively.

To ensure stable feature propagation across layers, the PCR leverages residual learning to integrate the adapted features $\mathbf{F}_a$ with $\mathbf{F}_l^{\mathrm{pc}\prime\prime}$. The outputs from both branches are subsequently combined to produce the point cloud features at the $l$-th layer, formulated as:
\begin{equation}
\mathbf{F}_l^{\mathrm{pc}} = \mathbf{F}_a + \text{FFN}(\mathbf{F}_l^{\mathrm{pc}\prime\prime}) + \mathbf{F}_l^{\mathrm{pc}\prime\prime}.
\end{equation}
Features extracted from transformer blocks at layers \(l \in \{3, 7, 11\}\) are concatenated to construct the holistic point representation \(\mathbf{F}^{\mathrm{pc}}\).
 
\subsection{Bidirectional Feature Alignment}
RGB images and 3D point clouds exhibit inherently distinct feature distributions, necessitating robust alignment to achieve effective multimodal integration. In contrast to traditional spatial alignment approaches, our method employs a mutual prediction mechanism to facilitate semantic alignment. This strategy enriches the textural detail of point cloud features while enhancing the spatial contextual awareness of RGB features, thereby effectively capturing complementary information across modalities.

Given the adaptive image and point cloud features $\mathbf{F}^{\mathrm{rgb}}$ and $\mathbf{F}^{\mathrm{pc}}$, we employ modality-specific predictors $\mathcal{H}^{\mathrm{rgb}}$ and $\mathcal{H}^{\mathrm{pc}}$, implemented as multi-layer perceptrons (MLPs), to bridge distributional gaps and capture complementary information. Specifically, $\mathcal{H}^{\mathrm{rgb}}$ predicts point cloud features $\hat{\mathbf{F}}^{\mathrm{pc}}$ from image features $\mathbf{F}^{\mathrm{rgb}}$, while $\mathcal{H}^{\mathrm{pc}}$ predicts image features $\hat{\mathbf{F}}^{\mathrm{rgb}}$ from point cloud features $\mathbf{F}^{\mathrm{pc}}$, formulated as:
\begin{equation}
\hat{\mathbf{F}}^{\mathrm{pc}} = \mathcal{H}^{\mathrm{rgb}}(\mathbf{F}^{\mathrm{rgb}}) \quad \hat{\mathbf{F}}^{\mathrm{rgb}} = \mathcal{H}^{\mathrm{pc}}(\mathbf{F}^{\mathrm{pc}})
\end{equation}
where $\hat{\mathbf{F}}^{\mathrm{rgb}}$ and $\hat{\mathbf{F}}^{\mathrm{pc}}$ denote the predicted RGB and point cloud features, respectively.

During training, $\mathcal{H}^{\mathrm{rgb}}$ and $\mathcal{H}^{\mathrm{pc}}$ are jointly optimized by minimizing the cosine distance between extracted and predicted features. To prevent feature collapse to a trivial solution and preserve modality-specific characteristics, we incorporate the stop-gradient operation. The loss function is defined as:
\begin{equation}
\mathcal{L} = \left(1 - \frac{\hat{\mathbf{F}}^{\mathrm{rgb}} \cdot \text{SG}(\mathbf{F}^{\mathrm{rgb}})}{\|\hat{\mathbf{F}}^{\mathrm{rgb}}\| \cdot \text{SG}(\|\mathbf{F}^{\mathrm{rgb}}\|)}\right) + \left(1 - \frac{\hat{\mathbf{F}}^{\mathrm{pc}} \cdot \text{SG}(\mathbf{F}^{\mathrm{pc}})}{\|\hat{\mathbf{F}}^{\mathrm{pc}}\| \cdot \text{SG}(\|\mathbf{F}^{\mathrm{pc}}\|)}\right),
\end{equation}
where $\text{SG}(\cdot)$ represents the stop-gradient operation.

The fine-tuned image and point cloud features are concatenated to form fused features, expressed as:
\begin{equation}
\mathbf{F}^{\mathrm{f}} = \text{concat}(\mathbf{F}^{\mathrm{rgb}}, \mathbf{F}^{\mathrm{pc}}).
\end{equation}

\subsection{Anomaly Detection}
To fully exploit both individual and fused modality information, we adopt the strategy from M3DM \citep{wang2023multimodal}, utilizing three memory banks—$\mathcal{M}^{\mathrm{rgb}}$, $\mathcal{M}^{\mathrm{pc}}$, and $\mathcal{M}^{\mathrm{f}}$ to store RGB, point cloud, and fused features, respectively. Anomaly scores are calculated using functions $\varphi$ and $\psi$ \citep{roth2022towards}:
\begin{alignat}{1}
& \varphi(\mathcal{M}, \mathbf{F}) = \eta \|\mathbf{F}^*(x,y) - m^*\|_2, \\
& \psi(\mathcal{M}, \mathbf{F}) = \left\{ \min_{m \in \mathcal{M}} \|\mathbf{F}^*(x,y) - m\|_2 \mid \mathbf{F}^*(x,y) \in \mathbf{F} \right\}, \\
& \mathbf{F}^*(x,y), m^* = \arg\max_{\mathbf{F}(x,y) \in \mathbf{F}} \arg\min_{m \in \mathcal{M}} \|\mathbf{F}(x,y) - m\|_2,
\end{alignat}
where $\eta$ is a re-weighting parameter, $\mathcal{M} \in \{\mathcal{M}^{\mathrm{rgb}}, \mathcal{M}^{\mathrm{pc}}, \mathcal{M}^{\mathrm{f}}\}$, and $\mathbf{F} \in \{\mathbf{F}^{\mathrm{rgb}}, \mathbf{F}^{\mathrm{pc}}, \mathbf{F}^{\mathrm{f}}\}$.

To integrate the anomaly scores derived from RGB, point cloud, and fused features, we employ two trainable One-Class Support Vector Machines (OCSVMs) \citep{scholkopf2001estimating}, denoted as $\mathcal{D}_i$ and $\mathcal{D}_p$, to balance the contributions of each modality. These generate the final Image-level anomaly score $S_i$ and pixel-level anomaly score $S_p$, respectively, formulated as:
\begin{flalign}
& S_i = \mathcal{D}_i(\varphi(\mathcal{M}^{\mathrm{rgb}}, \mathbf{F}^{\mathrm{rgb}}), \varphi(\mathcal{M}^{\mathrm{pc}}, \mathbf{F}^{\mathrm{pc}}), \varphi(\mathcal{M}^{\mathrm{f}}, \mathbf{F}^{\mathrm{f}})), \\
& S_p = \mathcal{D}_p(\psi(\mathcal{M}^{\mathrm{rgb}}, \mathbf{F}^{\mathrm{rgb}}), \psi(\mathcal{M}^{\mathrm{pc}}, \mathbf{F}^{\mathrm{pc}}), \psi(\mathcal{M}^{\mathrm{f}}, \mathbf{F}^{\mathrm{f}})).
\end{flalign}
\section{Experiments and analyses}\label{sec:experiment}
\subsection{Datasets and Evaluation Metrics}

\noindent\textbf{Datasets.} 
This study evaluates the proposed methods on two MAD datasets. MVTec 3D-AD ~\citep{bergmann2021mvtec} comprises 2,656 training, 294 validation, and 1,197 test samples. The training and validation sets consist exclusively of normal samples, while the test set includes 249 normal and 948 anomalous samples. Unlike traditional 2D image datasets, MVTec 3D-AD provides 4,147 scans captured by a high-resolution industrial 3D sensor, capturing both RGB images and 3D point cloud data concurrently. Eyecandies ~\citep{bonfiglioli2022eyecandies} contains 10,000 normal samples for training and 500 test samples, of which 250 exhibit complex rendered anomalies. Both datasets are leveraged to enhance anomaly detection by integrating 2D and 3D modalities.

\noindent\textbf{Data Preprocessing.} 
To optimize model performance on MAD, this study applies preprocessing techniques following established methods~\citep{wang2023multimodal,horwitz2023back}. The RANSAC algorithm~\citep{fischler1981random} is employed to estimate the background plane, removing points within 0.005 units of this plane. Corresponding RGB pixels are set to zero. Both the position tensor and RGB image are resized to $224 \times 224$ to meet the input requirements of the feature extractor.

\noindent\textbf{Implementation Details.} 
The proposed framework utilizes the pre-trained DINO ViT-B/8 model~\citep{caron2021emerging}, trained on ImageNet~\citep{krizhevsky2017imagenet}, and the Point-MAE model~\citep{zhao2021point}, trained on ShapeNet~\citep{shapenet}, as the backbones for the AIT and APT modules, respectively. AIT processes $224 \times 224$ RGB images, producing $28 \times 28 \times 768$ feature maps, which are upsampled to $56 \times 56 \times 768$ for fusion with point cloud features. APT extracts features from layers \{3, 7, 11\}, yielding a $56 \times 56 \times 1152$ feature vector. \jyx{For model optimization, both AIT and APT are fine-tuned using the Adam optimizer with a learning rate of $10^{-4}$ and a batch size of 8 over 50 epochs.} Additionally, two One-Class Support Vector Machines (OCSVMs), $\mathcal{D}_a$ and $\mathcal{D}_p$, are trained using Stochastic Gradient Descent (SGD) with a learning rate of \num{1e-4} for 1,000 steps per class. The implementation is executed using PyTorch v1.12.0 on an NVIDIA GeForce RTX 3090 Ti GPU and an Intel i9 @ 3.00GHz CPU.

\noindent\textbf{Evaluation Metrics.} 
Image-level anomaly detection performance is assessed using the Area Under the Receiver Operating Characteristic curve (I-AUROC), with higher values indicating superior detection capabilities. For anomaly localization, Pixel-level AUROC (P-AUROC) and Area Under the Per-Region Overlap (AUPRO) metrics are employed. AUPRO measures the average relative overlap between binary predictions and ground-truth connected components. While a False Positive Rate (FPR) threshold of 0.3 is commonly used for AUPRO, industrial applications often require stricter criteria to minimize false positives. Thus, this study reports AUPRO at both 0.3 (AUPRO@30\%) and 0.01 (AUPRO@1\%) thresholds.

\subsection{Anomaly Detection Results}

\noindent\textbf{Comparison with Baseline Anomaly Detection Methods.} 
The proposed BAT is benchmarked against leading MAD methods on both MVTec 3D-AD and Eyecandies. As presented in Table~\ref{first}, BAT achieves a 2.5\% improvement in I-AUROC over M3DM for anomaly detection on MVTec 3D-AD, with superior performance in eight out of ten anomaly detection categories. Notably, BAT demonstrates significant improvements in the potato and tire categories, with I-AUROC gains of 7.2\% and 12.0\%, respectively, compared to M3DM. Compared to the prior SOTA method, UCF, BAT attains a 0.7\% I-AUROC improvement while achieving optimal performance in P-AUROC and AUPRO@30\% for localization tasks. Under a stricter FPR threshold, as detailed in Table~\ref{second}, BAT maintains superior performance, validating its efficacy for high-precision industrial anomaly detection. 
Furthermore, as evidenced in Tables~\ref{third} and~\ref{four}, BAT matches M3DM's I-AUROC performance on the Eyecandies dataset while demonstrating significant advantages in localization, outperforming competing methods in P-AUROC, AUPRO@30\%, and AUPRO@1\%.

\begin{table*}[th]
  \centering
  \caption{Performance of anomaly detection and localization on MVTec 3D-AD dataset measured by I-AUROC, P-AUROC, and AUPRO@30\% scores. Best results are in \textbf{bold}.}
  \label{first}
  \resizebox{\textwidth}{!}{
    \begin{tabular}{ccccccccccccc}
      \toprule[1.5pt] 
      Metric     & Method    & \multicolumn{1}{l}{Bagel} & \multicolumn{1}{l}{Cable Gland} & \multicolumn{1}{l}{Carrot} & \multicolumn{1}{l}{Cookie} & \multicolumn{1}{l}{Dowel} & \multicolumn{1}{l}{Foam} & \multicolumn{1}{l}{Peach} & \multicolumn{1}{l}{Potato} & \multicolumn{1}{l}{Rope} & Tire           & Mean           \\
      \midrule
      \multirow{6}{*}{I-AUROC}
                       & BTF  \citep{horwitz2023back}     & 0.918                     & 0.748                           & 0.967                      & 0.883                      & 0.932                     & 0.582                    & 0.896                     & 0.912                      & 0.921                    & 0.886          & 0.865          \\
                       & AST    \citep{AST}   & 0.983                     & 0.873                           & 0.976                      & 0.971                      & 0.932                     & 0.885                    & 0.974                     & \textbf{0.981}                      & \textbf{1.000}                    & 0.797          & 0.937          \\
                       & M3DM  \citep{wang2023multimodal}    & 0.994                     &  \textbf{0.909}                           & 0.972                      & 0.976                      & 0.960                     & \textbf{0.942}                    & 0.973                     & 0.899                      & 0.972                    & 0.850          & 0.945          \\
                       & CFM   \citep{costanzino2024multimodal}    & 0.994                     & 0.888                           & 0.984                      & 0.993                      &0.980            & 0.888                    & 0.941                     & 0.943                      & 0.980                    & 0.953 & 0.954         \\ 
                      & UCF   \citep{UCF}    & \textbf{0.999}                     & 0.888                           & 0.978                      & 0.995                      & \textbf{0.997}            & 0.909                    & 0.984                     & 0.964                      & 0.985                    & 0.931 & 0.963         \\ 
                       & \cellcolor{gray!20}BAT     & \cellcolor{gray!20}0.996            &\cellcolor{gray!20}0.896                  & \cellcolor{gray!20}\textbf{0.992}             & \cellcolor{gray!20}\textbf{0.997}             & \cellcolor{gray!20}0.976                     & \cellcolor{gray!20}0.914           & \cellcolor{gray!20}\textbf{0.985}            & \cellcolor{gray!20}0.971             & \cellcolor{gray!20}0.998           & \cellcolor{gray!20}\textbf{0.970}          & \cellcolor{gray!20}\textbf{0.970}  \\
      \midrule
      \multirow{5}{*}{P-AUROC} 
                       & BTF   \citep{horwitz2023back}     & 0.996                     & 0.992                           & 0.997                      & \textbf{0.994}                      & 0.981                     & 0.974                    & 0.996                     & 0.998                      & 0.994                    & 0.995          & 0.992          \\
                       & AST   \citep{AST}    & -                     & -                           & -                      & -                      & -                     & -                    & -                     & -                      & -                    & -          & 0.976          \\
                       & M3DM   \citep{wang2023multimodal}   & 0.995                     & \textbf{0.993}                           & 0.997                      & 0.985                      & 0.985                     & 0.984                    & 0.996                     & 0.994                      & 0.997                    & 0.996          & 0.992          \\                         
                       & CFM   \citep{costanzino2024multimodal}    & \textbf{0.997}            & 0.992                  & \textbf{0.999}             & 0.972                      & 0.987            & 0.993                    & \textbf{0.998}            & \textbf{0.999}             & \textbf{0.998}           & \textbf{0.998}          & 0.993          \\
                       & \cellcolor{gray!20}BAT       & \cellcolor{gray!20}\textbf{0.997}            & \cellcolor{gray!20}0.992                  & \cellcolor{gray!20}0.998                      & \cellcolor{gray!20}0.984             & \cellcolor{gray!20}\textbf{0.997}                     & \cellcolor{gray!20}\textbf{0.998}           & \cellcolor{gray!20}\textbf{0.998}            & \cellcolor{gray!20}0.998             & \cellcolor{gray!20}0.996           & \cellcolor{gray!20}0.994 & \cellcolor{gray!20}\textbf{0.995} \\
      \midrule
      \multirow{6}{*}{AUPRO@30\%}   
                       & BTF \citep{horwitz2023back} &0.976&0.969&0.979&\textbf{0.973}&0.933&0.888&0.975&0.981&0.950&0.971&0.959\\
                       & AST \citep{AST}&0.970&0.947&0.981&0.939&0.913&0.906&0.979&0.982&0.889&0.940&0.944\\
                       & M3DM   \citep{wang2023multimodal}   & 0.970                     & 0.971                           & 0.979                      & 0.950                      & 0.941                     & 0.932                    & 0.977                     & 0.971                      & 0.971                    & 0.975          & 0.964          \\
                       & CFM  \citep{costanzino2024multimodal}     & 0.979                     & \textbf{0.972}                  & \textbf{0.982}             & 0.945                      & 0.950            & 0.968                    & 0.980                     & 0.982                      & 0.975                    & \textbf{0.981}          & 0.971          \\
                         & UCF   \citep{UCF}    & \textbf{0.981}                     & 0.977                           & \textbf{0.982}                      & 0.951                      & 0.961            & 0.973                    & \textbf{0.983}                     & \textbf{0.983}                      & \textbf{0.981}                    & 0.980 & 0.975         \\ 
                       & \cellcolor{gray!20}BAT      & \cellcolor{gray!20}\textbf{0.981}            & \cellcolor{gray!20}0.970                           & \cellcolor{gray!20}\textbf{0.982}                      & \cellcolor{gray!20}0.960             & \cellcolor{gray!20}\textbf{0.977}                     & \cellcolor{gray!20}\textbf{0.981}           & \cellcolor{gray!20}0.982            & \cellcolor{gray!20}0.982             & \cellcolor{gray!20}0.977           & \cellcolor{gray!20}0.965 & \cellcolor{gray!20}\textbf{0.976}\\
      \bottomrule[1.5pt]  
    \end{tabular}
  }
\end{table*}
\begin{table*}[th]
  \centering
  \scriptsize
  \caption{Performance of anomaly localization on MVTec 3D-AD dataset measured by AUPRO@1\% scores. Best results are in \textbf{bold}.}
  \label{second}
\scriptsize
  \resizebox{\textwidth}{!}{ 
    \sisetup{table-format=1.3}
    \begin{tabular}{l *{11}{S}}
      \toprule
      Method & {Bagel} & {Cable Gland} & {Carrot} & {Cookie} & {Dowel} & {Foam} & {Peach} & {Potato} & {Rope} & {Tire} & {Mean} \\
      \midrule
      AST & 0.388 & 0.322 & 0.470 & 0.411 & 0.328 & 0.275 & 0.474 & 0.487 & 0.360 & 0.474 & 0.398 \\
      M3DM & 0.414 & 0.395 & 0.447 & 0.318 & \textbf{0.422} & 0.335 & 0.444 & 0.351 & 0.416 & 0.398 & 0.394 \\
      CFM & 0.459 & \textbf{0.431} & \textbf{0.485} & \textbf{0.469} & 0.394 & 0.413 & 0.468 & 0.487 & \textbf{0.464} & 0.476 & 0.455 \\
      \rowcolor{gray!20} 
      BAT & \textbf{0.492} & \textbf{0.431} & \textbf{0.485} & \textbf{0.469} & 0.354 & \textbf{0.490} & \textbf{0.495} & \textbf{0.493} & \textbf{0.464} & \textbf{0.493} & \textbf{0.465} \\
      \bottomrule
    \end{tabular}
  }
\end{table*}
\begin{table*}[tb]
  \centering
          \caption{Performance of anomaly detection and localization on Eyecandies dataset measured by I-AUROC, P-AUROC, and AUPRO@30\% scores. Best results are in \textbf{bold}.}
\label{third}
  \resizebox{\textwidth}{!}{
\begin{tabular}{ccccccccccccc}
\toprule[1.5pt]
Metric &Method&Can. C.&Cho. C.&Cho. P.&Conf.&Gum. B.&Haz. T.& Lic.S.&Lollip.&Marsh.&Pep. C.&Mean\\
\midrule  
\multirow{6}*{I-AUROC}&
RGB-D  \citep{bonfiglioli2022eyecandies}&0.529&0.861&0.739&0.752&0.594&0.498&0.679&0.651& 0.838&0.750&0.689\\
&RGB-cD-n  \citep{bonfiglioli2022eyecandies}&0.596&0.843&0.819&0.846&0.833&0.550&0.750&0.846 &0.940&0.848&0.787\\
&AST \citep{AST}&0.574&0.747&0.747&0.889&0.596&0.617&0.816&0.841 &0.987&0.987&0.780\\
&M3DM \citep{wang2023multimodal}&0.624&0.958&\textbf{0.958}&\textbf{1.000}&0.886&0.758&\textbf{0.949}&0.836&\textbf{1.000}&\textbf{1.000}&\textbf{0.897}\\
&CFM\citep{costanzino2024multimodal} &\textbf{0.680}&0.931&0.952&0.880&0.865&\textbf{0.782}&0.917&0.840&0.998 &0.962&0.881\\
&\cellcolor{gray!20}BAT \cellcolor{gray!20}&\cellcolor{gray!20}0.605&\cellcolor{gray!20}\textbf{0.970}&\cellcolor{gray!20}0.950&\cellcolor{gray!20}0.995&\cellcolor{gray!20}\textbf{0.902}&\cellcolor{gray!20}0.654&\cellcolor{gray!20}0.930&\cellcolor{gray!20}\textbf{0.853}&\cellcolor{gray!20}0.998&\cellcolor{gray!20}\textbf{1.000}&\cellcolor{gray!20}0.886
\\
\midrule
\multirow{6}*{P-AUROC}&
RGB-D \citep{bonfiglioli2022eyecandies}&0.973&0.927&0.958&0.945&0.929&0.806&0.827&0.977& 0.931&0.928&0.920\\
&RGB-cD-n  \citep{bonfiglioli2022eyecandies}&0.980&0.979&\textbf{0.982}&0.978&0.951&0.853&0.971&0.978& 0.985&0.967&0.962\\
&AST \citep{AST}&0.763&0.960&0.911&0.969&0.788&0.837&0.918&0.924 &0.983&0.968&0.902\\
&M3DM \citep{wang2023multimodal}&0.974&\textbf{0.987}&0.962&\textbf{0.998}&0.966&0.941&0.973&0.984 &0.996&0.985&0.977\\
&CFM\citep{costanzino2024multimodal} &\textbf{0.983}&0.982&0.964&0.989&0.949&\textbf{0.946}&0.969&0.980&0.995&0.987&0.974\\
&\cellcolor{gray!20}BAT &\cellcolor{gray!20}0.973&\cellcolor{gray!20}0.984&\cellcolor{gray!20}0.969&\cellcolor{gray!20}\textbf{0.998}&\cellcolor{gray!20}\textbf{0.976}&\cellcolor{gray!20}0.943&\cellcolor{gray!20}\textbf{0.977}&\cellcolor{gray!20}\textbf{0.986}&\cellcolor{gray!20}\textbf{0.998}&\cellcolor{gray!20}\textbf{0.996}&\cellcolor{gray!20}\textbf{0.980}
\\
\midrule
\multirow{4}*{AUPRO@30\%}&AST \citep{AST}&0.514&0.835&0.714&0.905&0.587&0.590&0.736&0.769&0.918&0.878&0.744\\
&M3DM \citep{wang2023multimodal}&0.906&\textbf{0.923}&0.803&\textbf{0.983}&0.855&0.688&0.880&0.906& 0.966&0.955&0.882\\
&CFM\citep{costanzino2024multimodal}&\textbf{0.942}&0.902&0.831& 0.965&\textbf{0.875}&\textbf{0.762}&0.791&\textbf{0.913}&0.939&0.949&0.887\\
&\cellcolor{gray!20}BAT&\cellcolor{gray!20}0.910&\cellcolor{gray!20}0.912&\cellcolor{gray!20}\textbf{0.834}	&\cellcolor{gray!20}0.982&\cellcolor{gray!20}\textbf{0.875}&\cellcolor{gray!20}0.684&\cellcolor{gray!20}\textbf{0.893}&\cellcolor{gray!20}0.888&\cellcolor{gray!20}\textbf{0.974}&\cellcolor{gray!20}\textbf{0.975}&\cellcolor{gray!20}\textbf{0.893}
\\
\bottomrule[1.5pt]
\end{tabular}
}
\end{table*}
\begin{table*}[tbp]
  \centering
  \scriptsize
  \caption{Performance of anomaly localization on Eyecandies dataset measured by AUPRO@1\% scores. Best results are in \textbf{bold}.}
  \label{four}
  \scriptsize
  \resizebox{\textwidth}{!}{ 
    \sisetup{table-format=1.3}
    \begin{tabular}{l *{10}{S} S[table-format=1.3]}
      \toprule
      Method & {Can. C.} & {Cho. C.} & {Cho. P.} & {Conf.} & {Gum. B.} & {Haz. T.} & {Lic. S.} & {Lollip.} & {Marsh.} & {Pep. C.} & {Mean} \\
      \midrule
      AST & 0.035 & 0.230 & 0.129 & 0.234 & 0.092 & 0.069 & 0.139 & 0.090 & 0.255 & 0.224 & 0.149 \\
      M3DM & 0.166 & 0.388 & 0.329 & 0.486 & 0.315 & 0.131 & 0.323 & 0.258 & 0.462 & 0.454 & 0.331 \\
      CFM & \textbf{0.229} & 0.397 & \textbf{0.345} & 0.389 & \textbf{0.353} & \textbf{0.188} & 0.333 & 0.236 & 0.455 & 0.428 & 0.335 \\
      \rowcolor{gray!20} 
      BAT & 0.192 & \textbf{0.403} & \textbf{0.345} & \textbf{0.491} & \textbf{0.353} & \textbf{0.188} & \textbf{0.345} & \textbf{0.266} & \textbf{0.467} & \textbf{0.458} & \textbf{0.348} \\
      \bottomrule
    \end{tabular}
  }
\end{table*}

\noindent\textbf{Comparison with Few-Shot Anomaly Detection Methods.} 
Given the challenges of collecting large-scale datasets in real-world scenarios, effective anomaly detection models must accurately model normal data distributions under limited samples~\citep{costanzino2024multimodal}. To evaluate BAT’s proficiency in this context, few-shot experiments are conducted using 5, 10, and 50 randomly selected image-point cloud pairs per category from MVTec 3D-AD. As demonstrated in Table~\ref{few_shot}, BAT consistently surpasses top-performing MAD methods, including BTF, AST, and M3DM, in both anomaly detection and localization performance across all few-shot settings.

\begin{table*}[htb]
  \begin{center}
        \caption{Performance of few-shot anomaly detection and localization on MVTec 3D-AD dataset measured by I-AUROC, P-AUROC, and AUPRO@30\% scores. Best results are in \textbf{bold}.}
\label{few_shot}
  \resizebox{\textwidth}{!}{
\begin{tabular}{ccccccccccccc}
\toprule[1.5pt] 
\multirow{2}*{Method}&5-shot&10-shot&50-shot&Full&5-shot&10-shot&50-shot&Full&5-shot&10-shot&50-shot&Full\\
\cmidrule(l){2-5} \cmidrule(l){6-9} \cmidrule(l){10-13}
&\multicolumn{4}{c}{I-AUROC}&\multicolumn{4}{c}{P-AUROC}&\multicolumn{4}{c}{AUPRO}\\
\midrule 
BTF&0.671&0.695&0.806&0.865&0.980&0.983&0.989&0.992&0.920&0.928&0.947&0.959\\
AST &0.680&0.689&0.794&0.937&0.950&0.946&0.974&0.976&0.903&0.835&0.929&0.944\\
M3DM &0.796&0.821&0.903&0.945&0.984&0.986&0.989&0.992&0.927&0.939&0.952&0.964\\
\rowcolor{gray!20}BAT &\textbf{0.798}&\textbf{0.842}&\textbf{0.908}&\textbf{0.966}&\textbf{0.987}&\textbf{0.989}&\textbf{0.991}&\textbf{0.995}&\textbf{0.950}&\textbf{0.956}&\textbf{0.959}&\textbf{0.973}\\
\bottomrule[1.5pt]
\end{tabular}
}
\end{center}
\end{table*}

\subsection{Qualitative Analysis} 
\begin{figure*}[htbp]
\centering
\includegraphics[width=0.95\columnwidth]{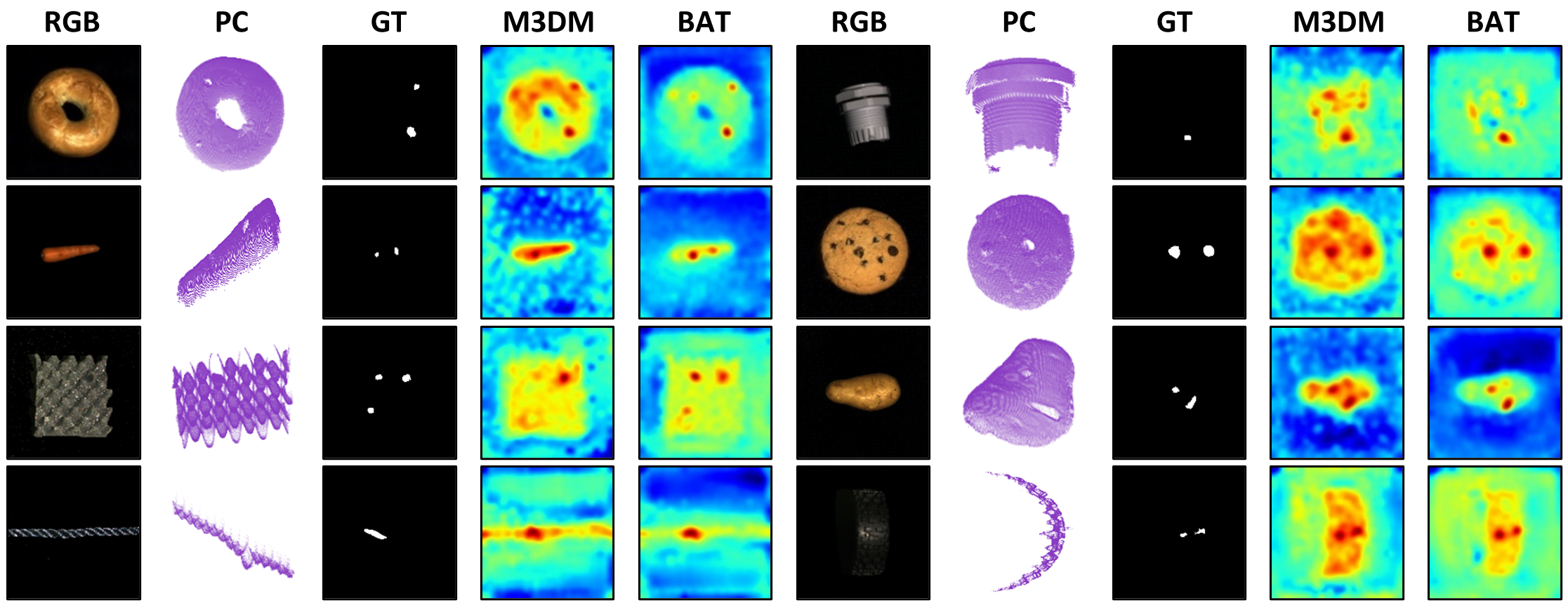}
\caption{\textbf{Qualitative comparison of multimodal anomaly detection performance.} Unlike M3DM, which produces broader and less precise anomaly maps often highlighting object outlines, BAT generates compact and accurate anomaly maps with minimal disruption, showcasing superior localization performance.}
\label{fig:five}
\end{figure*}

Figure~\ref{fig:five} presents qualitative comparisons between BAT and M3DM on MVTec 3D-AD. M3DM tends to produce anomaly maps with broader, imprecise anomaly regions, often emphasizing object outlines. In contrast, BAT generates compact, accurate anomaly maps with minimal disruption, demonstrating superior localization capability.

\subsection{Ablation Study} 
Ablation studies are performed to evaluate the individual contributions of BAT’s constituent components, with detailed findings elaborated in the subsequent sections.

\noindent\textbf{Analysis on AIT.} 
\jyx{The AIT module is designed as a parameter-efficient component to refine generic image features for domain-specific anomaly detection. As quantified in Table~\ref{five}, the integration of AIT yields a $1.9\%$ improvement in I-AUROC over the frozen image backbone without an adaptor, and this performance gain is achieved with only a minimal parameter increase from 327MB to 343MB.}
Furthermore, a comprehensive evaluation of various image feature adaptors underscores the superiority of convolutional layers over linear layers in discerning intricate semantic patterns. The incorporation of a CoordConv layer further augments AIT's performance by enhancing its spatial contextual awareness, thereby yielding optimal results in MAD tasks.

\begin{table}[htb]
\centering
\caption{Performance comparison of the AIT module across different image adaptor structures. Best results are in \textbf{bold}.}
\label{five}
\setlength{\tabcolsep}{10pt}
\renewcommand{\arraystretch}{1}
\begin{tabular}{cccc}  
\toprule[1.5pt]
Structure & I-AUROC & P-AUROC & AUPRO@30\% \\
\midrule
Image Backbone & 0.850 & \textbf{0.987} & 0.942 \\
Linear Projection & 0.843 & 0.986 & 0.943 \\
MLP & 0.537 & 0.908 & 0.722 \\
Convolution & 0.863 & 0.986 & \textbf{0.944} \\
\rowcolor{gray!20}
AIT & \textbf{0.869} & 0.986 & \textbf{0.944} \\
\bottomrule[1.5pt]
\end{tabular}
\end{table}
\begin{table}[htb]
\centering
\caption{Performance comparison of the APT module against the point backbone and full fine-tuning approaches. Best results are in \textbf{bold}.}
\label{six}
\setlength{\tabcolsep}{10pt}
\renewcommand{\arraystretch}{1}
\begin{tabular}{cccc}  
\toprule[1.5pt]
Structure & I-AUROC & P-AUROC & AUPRO@30\% \\
\midrule
Point Bakcbone & 0.875 & 0.970 & 0.906 \\
Full-Parameter Training & 0.908 & \textbf{0.980} & 0.932 \\
\rowcolor{gray!20}
APT & \textbf{0.921} & \textbf{0.980} & \textbf{0.933} \\
\bottomrule[1.5pt]
\end{tabular}
\end{table}
\begin{table}[t]
\centering
\caption{Performance comparison of different 2D and 3D feature alignment strategies. Best results are in \textbf{bold}.}
\label{eight}
\setlength{\tabcolsep}{10pt}
\renewcommand{\arraystretch}{1}
\begin{tabular}{cccc}
\toprule[1.5pt] 
Alignment               & I-AUROC        & P-AUROC        & AUPRO@30\%         
\\ \midrule  
Concatenation          & 0.864          & 0.986          & 0.945          \\
Spatial Alignment & 0.945	&0.992	&0.971          \\
2D-to-3D Alignment & 0.922          & 0.988          & 0.959          \\
3D-to-2D Alignment & 0.945          & 0.991          & 0.966          \\
\rowcolor{gray!20} Semantic Alignment (BFA)  & \textbf{0.970} & \textbf{0.995} & \textbf{0.976}\\
\bottomrule[1.5pt]
\end{tabular}
\end{table}
\noindent\textbf{Analysis on APT.} 
The effectiveness of the APT module in anomaly detection is demonstrated in Table~\ref{six}. \jyx{The point backbone, an unadapted pre-trained network, yields the lowest performance across all evaluated metrics, indicating its limited adaptability to specialized industrial scenarios.} Fine-tuning the entire set of transformer block parameters in the pre-trained network results in suboptimal outcomes, primarily due to the risks of catastrophic forgetting, which also imposes significant computational costs. \jyx{In contrast, the APT module achieves substantial enhancements, with a 4.6\% improvement in I-AUROC and a 2.7\% increase in AUPRO compared to the baseline point backbone, while introducing only an additional 2MB of tunable parameters (83MB $\rightarrow$ 85MB). These results underscore APT’s capacity to significantly improve anomaly localization accuracy while effectively mitigating the forgetting risks.}

\begin{figure*}[t]
\centering
\includegraphics[width=0.6\columnwidth]{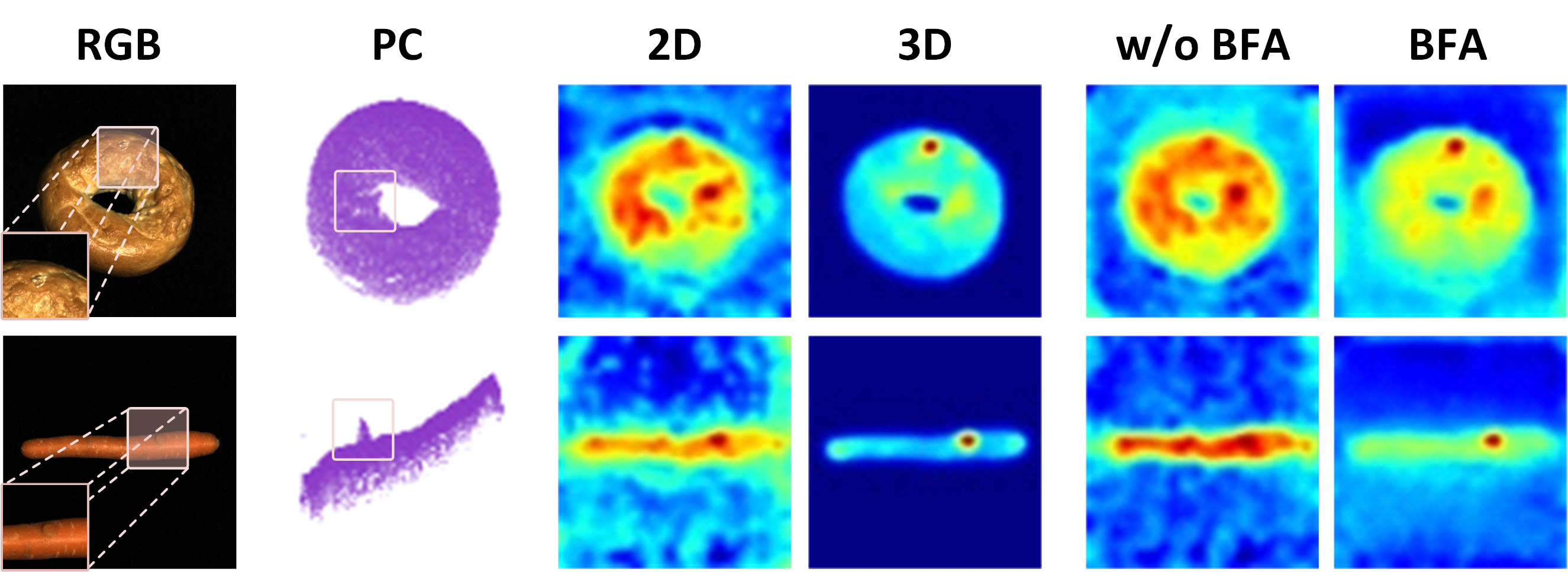}
\caption{\textbf{Visualization of multimodal anomaly localization results with and without the BFA module.} BFA reduces feature interference, significantly enhancing localization accuracy.}

\label{fig:BFA}
\end{figure*}

\noindent\textbf{Analysis on BFA.} 
\jyx{We validate the integration effect of the BFA module within the full multimodal framework. As shown in Table~\ref{eight}, by incorporating the AIT and APT together with the BFA module, the BAT framework—equipped with a memory bank and OCSVM for final anomaly detection—achieves the best overall performance (0.970/0.995/0.976). For comparison, direct concatenation of unaligned 2D and 3D features yields the least effective anomaly detection performance. Moreover, a configuration employing spatial alignment but using only frozen backbones (without AIT and APT) while retaining memory bank and OCSVM attains 0.945/0.992/0.971. Although this result demonstrates the inherent effectiveness of the memory bank and OCSVM detection modules, the complete model comprehensively surpasses it, indicating that the performance ceiling of the memory bank and OCSVM is highly dependent on the quality of input features. The high-quality aligned features produced by AIT and APT under BFA's guidance enable these support modules to realize their full potential.
Furthermore, as illustrated in Figure~\ref{fig:BFA}, the BFA module leverages semantic alignment to effectively integrate complementary information from 2D RGB images and 3D point cloud modalities. This approach minimizes cross-modal feature interference in scenarios where anomalous textures in RGB images exhibit limited discernibility, achieving precise multimodal anomaly localization.}

\begin{figure*}[t]
\centering
\includegraphics[width=0.95\columnwidth]{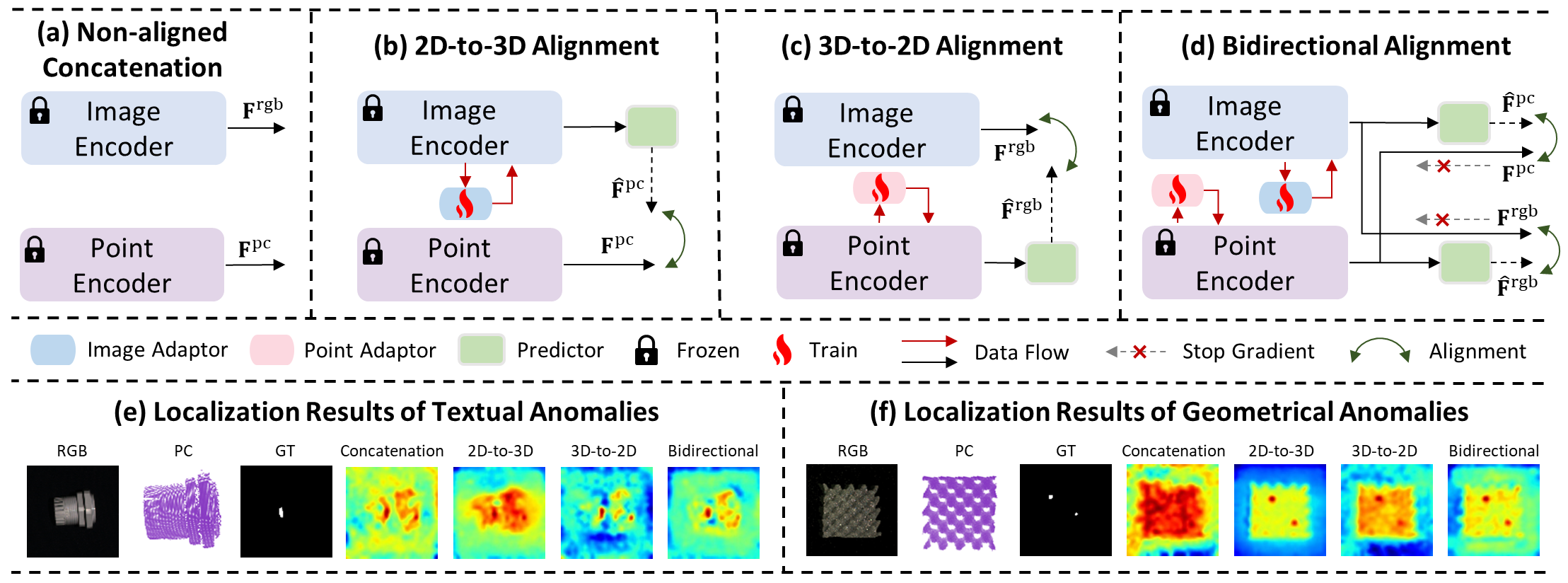}
\caption{\textbf{Visualization of different alignment strategies between 2D and 3D features.} (a)--(d) depict non-aligned concatenation, unidirectional 2D-to-3D alignment, unidirectional 3D-to-2D alignment, and bidirectional alignment, respectively. (e) and (f) showcase their effects on semantic and geometric anomaly localization, with BFA achieving the most accurate results.}
\label{fig:seven}
\end{figure*}

Beyond quantitative comparisons, we also present qualitative results of the four different alignment methods in Fig.~\ref{fig:seven}. (a) Direct concatenation of unaligned 2D and 3D features suffers from cross-modal distributional disparities, leading to imprecise anomaly localization. (b) Unidirectional 2D-to-3D alignment is predominantly influenced by 3D features, proving advantageous for detecting geometrical anomalies, such as protruding regions in foam, but less effective for semantic anomalies, such as threads in cable glands. (c) Conversely, unidirectional 3D-to-2D alignment prioritizes 2D features, which may limit its ability to capture geometric anomalies. (d) The BFA approach, leveraging bidirectional alignment, effectively unifies complementary 2D and 3D feature representations, resulting in superior precision for both semantic and geometric anomaly localization.

\begin{figure*}[t]
\centering
\includegraphics[width=0.55\columnwidth]{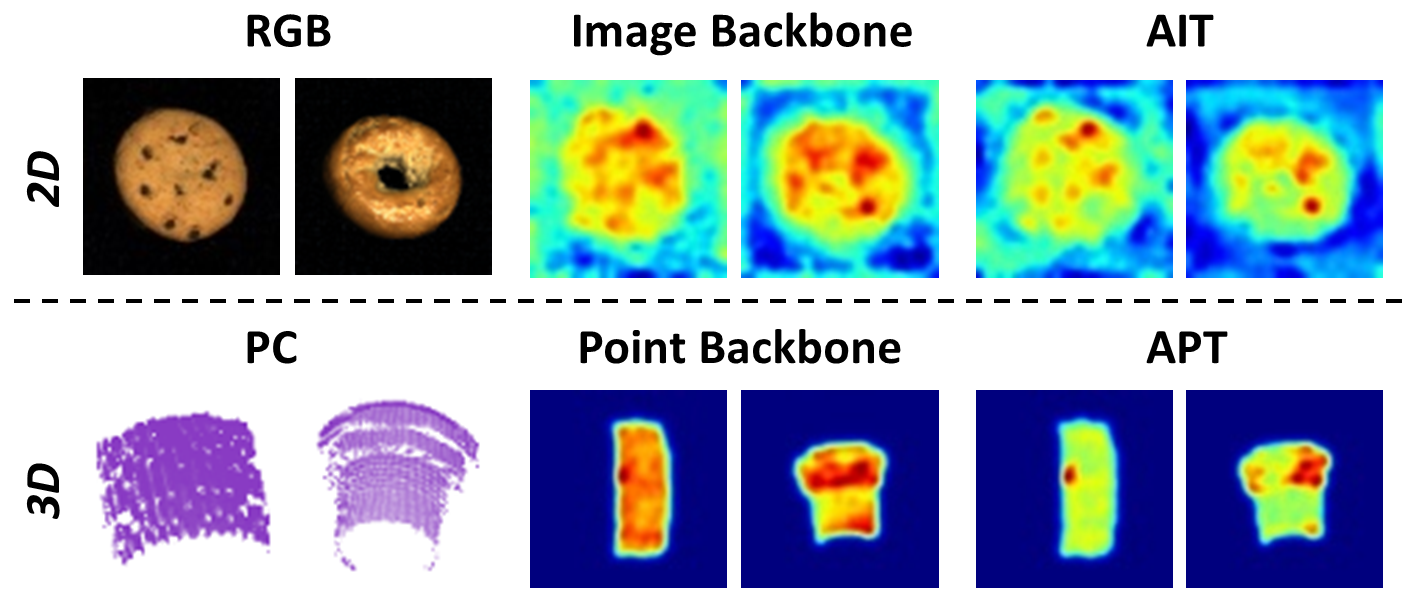}
\caption{\textbf{Visualization of 2D/3D single-modal anomaly detection results.} Compared to the image and point backbones, both AIT and APT achieve superior precision in anomaly localization.}
\label{fig:Single}
\end{figure*}

\noindent\textbf{Analysis on Single-modal Anomaly Detection.}
The proficiency of the BAT method in single-modal anomaly localization is showcased in Figure~\ref{fig:Single}. Within the 2D modality, the image backbone frequently misclassifies complex textures in normal regions as anomalies, whereas the AIT module within BAT ensures precise localization through enhanced geometric feature extraction during training. Likewise, in the 3D point cloud modality, the point backbone erroneously designates entire entities as anomalous, failing to achieve accurate localization. In contrast, BAT excels by precisely delineating anomalous regions. These experimental results further substantiate that our approach augments the extraction of modality-complementary features, refining the textural granularity of point cloud data while enhancing the spatial contextual awareness of RGB features.

\begin{figure}[t]
\centering
\includegraphics[width=0.5\columnwidth]{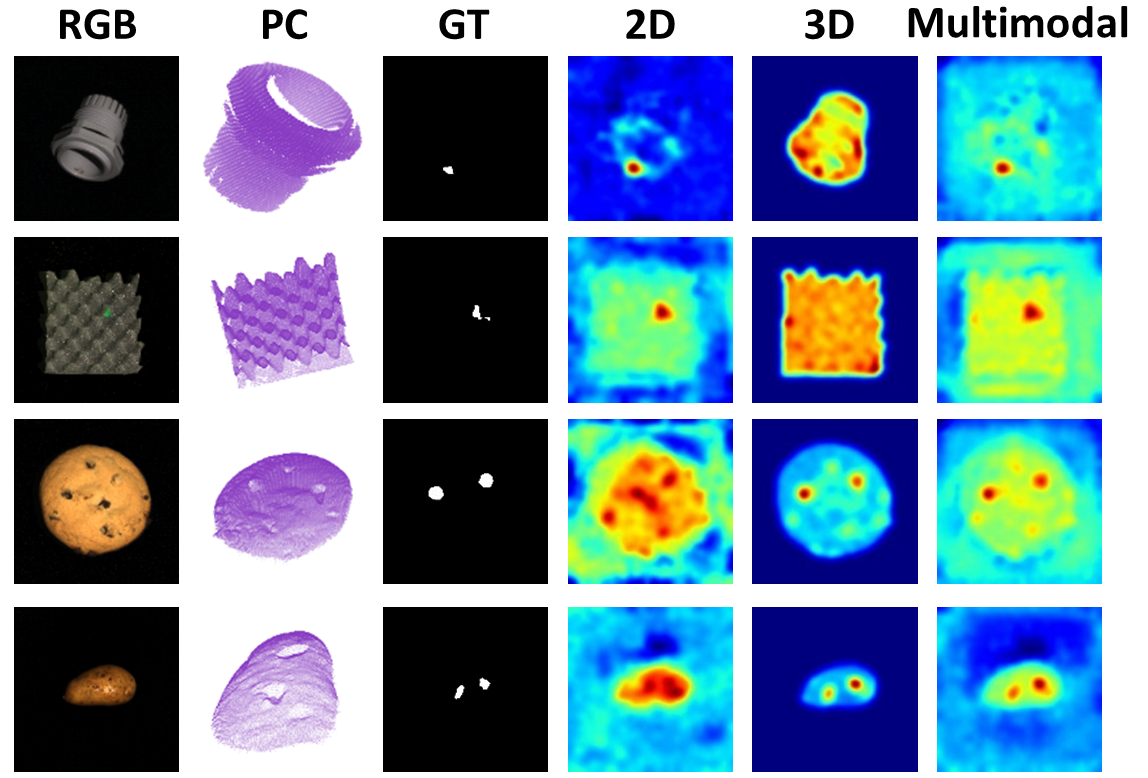}
\caption{\textbf{Visualization of multimodal anomaly detection results }. Anomaly localization under the multimodal setting demonstrates superior performance compared to the single-modal settings.}
\label{fig:eight}
\end{figure}

\noindent\textbf{Analysis on Multimodal Anomaly Detection.} 
Figure~\ref{fig:eight} evaluates the influence of multimodal data on anomaly localization. In contrast to single-modality methods, where the 2D modality excels at detecting texture variations and the 3D modality outperforms in identifying geometric anomalies, BAT multimodal approach synergistically combines the complementary strengths of both modalities. This integration significantly enhances localization accuracy, achieving superior performance over single-modality counterparts.

\jyx{To provide a comprehensive quantitative overview of all components discussed above, we summarize the complete ablation study results in Table~\ref{tab:ablation_modules}. The table consolidates the performance gains from each module and their integrated effect within the full BAT framework.}

\begin{table}[htb]
\centering
\caption{{Comprehensive ablation study results of single-modal adapters (AIT/APT) and multimodal bidirectional alignment within the BAT framework.}}
\label{tab:ablation_modules}
\setlength{\tabcolsep}{10pt}
\renewcommand{\arraystretch}{1}
\begin{tabular}{cccccc}  
\toprule[1.5pt]
Modality & Module & Module size & I-AUROC & P-AUROC & AUPRO \\
\midrule
2D & w/o AIT & 327MB & 0.850 & 0.987 & 0.942 \\
\rowcolor{gray!20}  
2D & w/ AIT & 343MB & 0.869 & 0.986 & 0.944 \\
3D & w/o APT & 83MB & 0.874 & 0.970 & 0.906 \\
\rowcolor{gray!20}  
3D & w/ APT & 85MB & 0.921 & 0.980 & 0.933 \\
\rowcolor{gray!20}  
Multimodal & w/ BFA & 484MB & \textbf{0.970} & \textbf{0.995} & \textbf{0.976} \\
\bottomrule[1.5pt]
\end{tabular}
\end{table}


\begin{figure}[t]
\centering
\includegraphics[width=0.5\columnwidth]{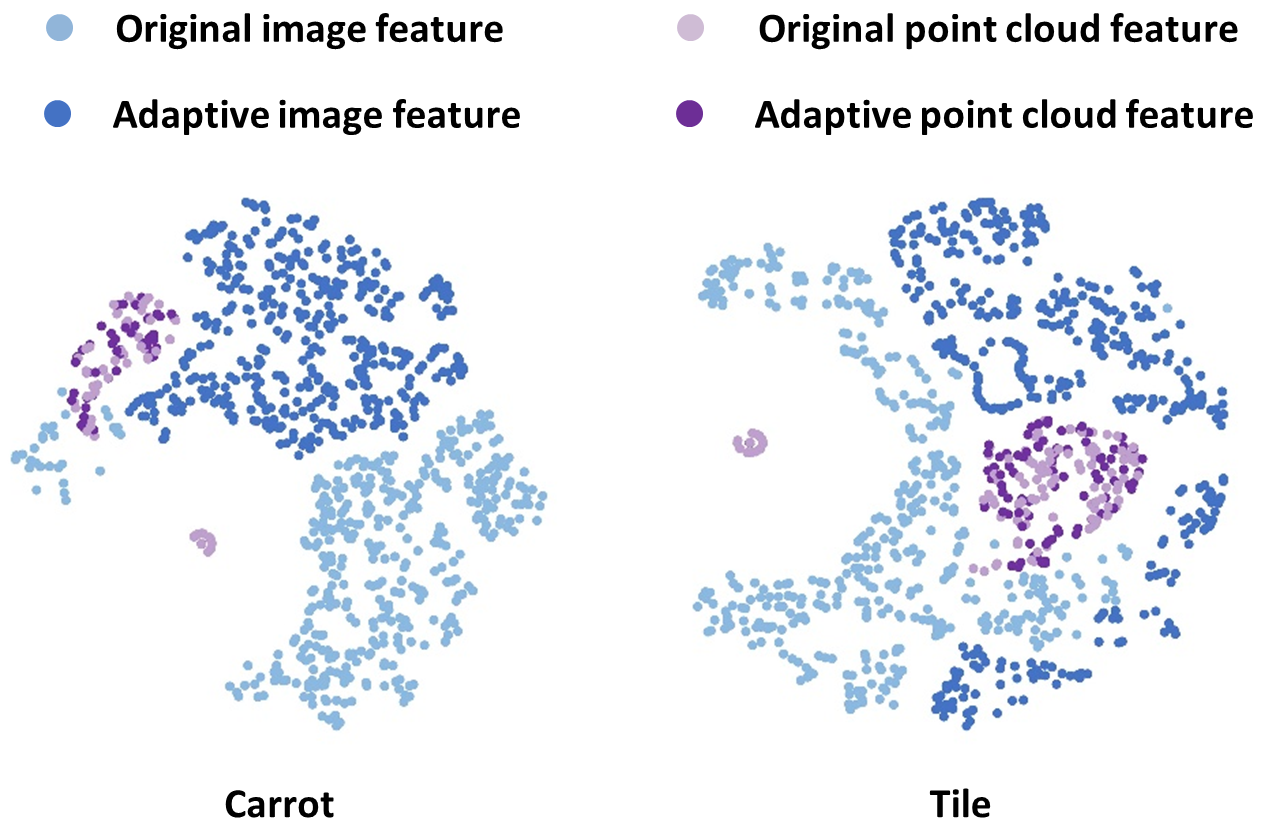}
\caption{\textbf{t-SNE visualization of marginal group distributions for “carrot” and “tile” objects from MVTec 3D-AD.}}
\label{fig:nine}
\end{figure}

\noindent\textbf{Analysis on Modality Feature Distribution.}
T-SNE visualization in Figure~\ref{fig:nine} illustrates the feature distribution of the MVTec 3D dataset. Original point cloud features exhibit a bifurcated distribution due to pooling operations at non-point and point cloud boundaries. In contrast, adaptive point cloud features align closely with adaptive image features, forming a compact distribution that reduces feature interference and enhances suitability for MAD task.

\noindent\textbf{Analysis on Different Backbones.} 
This section evaluates the impact of different backbone configurations for RGB images and point clouds, compared against the M3DM baseline. For RGB images, four extractor settings are analyzed: (1) ViT-B/8 supervised backbone pretrained on ImageNet 1K, (2) ViT-B/8 supervised backbone pretrained on ImageNet 21K, (3) ViT-S/8 self-supervised backbone pretrained via DINO~\citep{caron2021emerging}, and (4) ViT-B/8 self-supervised backbone pretrained via DINO. For point clouds, two self-supervised backbones are considered: (1) Point-Bert~\citep{yu2022point} and (2) Point-MAE~\citep{pang2022masked}, paired with the ViT-B/8 self-supervised RGB backbone.
Results in Table~\ref{backbones} demonstrate that BAT consistently outperforms M3DM across all configurations in both 2D and 3D modalities. Notably, smaller backbones pretrained with self-supervised methods outperform larger supervised pretrained backbones, likely due to their reduced parameter complexity and simpler architectures, which prevent overfitting and improve generalization. Additionally, Point-MAE surpasses Point-Bert, primarily due to its superior reconstruction of fine-grained point cloud details, which enhances the detection of subtle anomalies.

\begin{table}[hb!]
\centering
\caption{Performance comparison of different backbones. Best results are in \textbf{bold}.}
\label{backbones}
\setlength{\tabcolsep}{6pt}
\renewcommand{\arraystretch}{1}
\begin{tabular}{ccccc}
\toprule[1.5pt]  
                     & Method                & Backbone                & I-AUROC        & AUPRO@30\%          \\ \midrule 
\multirow{8}{*}{2D} & \multirow{4}{*}{M3DM} & Supervised ImageNet 1K  & 0.670          & 0.808          \\
                     &                       & Supervised ImageNet 21K & 0.662          & 0.837          \\
                     &                       & DINO ViT-S/8            & 0.819          & 0.937          \\
                     &                       & DINO ViT-B/8            & 0.850          & 0.942          \\\cmidrule(l){2-5}
                    & \multirow{4}{*}{BAT}  & Supervised ImageNet 1K  & 0.697          & 0.822          \\  
                     &                       & Supervised ImageNet 21K & 0.679          & 0.843          \\
                     &                       & DINO ViT-S/8            & 0.830          & 0.936          \\
                     &                       & \cellcolor{gray!20} DINO ViT-B/8            & \cellcolor{gray!20}\textbf{0.869} & \cellcolor{gray!20}\textbf{0.944} \\\midrule 
\multirow{4}{*}{3D}  & \multirow{2}{*}{M3DM} & Point-Bert              & 0.803          & 0.873          \\
                     &                       & Point-MAE               & 0.874          & 0.906          \\\cmidrule(l){2-5}
                     & \multirow{2}{*}{BAT}  & Point-Bert              & 0.895          & 0.931          \\  
                     &                       & \cellcolor{gray!20}Point-MAE               & \cellcolor{gray!20}\textbf{0.921} & \cellcolor{gray!20}\textbf{0.933}\\
 \bottomrule[1.5pt]
\end{tabular}
\end{table}

\jyx{\noindent\textbf{Limitations.} 
Although the proposed BAT framework achieves notable performance improvements in multimodal anomaly detection, several limitations persist. Specifically, BAT requires paired RGB and point cloud data during training, as its bidirectional alignment module relies on such paired samples to learn effective cross-modal correspondences. Nevertheless, collecting large-scale paired RGB-point cloud data remains challenging in practical industrial environments. To mitigate this limitation, one promising direction is to explore zero-shot multimodal alignment by pre-training the bidirectional alignment module on large-scale synthetic or publicly available multimodal datasets, and subsequently transferring the learned representations to target domains where paired data are scarce.}

\section{Conclusion and future work}\label{sec:conclusion}
This research proposes the Bidirectional Adaptive Transformers (BAT) framework, a novel method for multimodal anomaly detection (MAD) that constructs comprehensive and semantically aligned multimodal representations. BAT introduces a bidirectional feature alignment module that enables mutual feature prediction across modalities, thereby simultaneously enhancing the textural granularity of point cloud features and the spatial contextual awareness of RGB features. Furthermore, BAT integrates lightweight adapters within pre-trained transformer networks, facilitating efficient fine-tuning to preserve rich information essential for semantic alignment. Extensive experiments on the MVTec 3D-AD and Eyecandies validate BAT’s superior performance, surpassing existing MAD  methodologies.

\jyx{Although BAT makes notable advancements in integrating RGB and point cloud data for MAD, several avenues for further refinement exist.  Future research could investigate the incorporation of auxiliary modalities, such as linguistic or depth data, to further enrich the multimodal representation space. Moreover, extending BAT to encompass multi-illumination and multi-view RGB images alongside 3D point cloud data would be a valuable direction, as different lighting conditions and viewpoints can reveal complementary defect patterns, further improving detection robustness in complex scenarios.}

\section*{Acknowledgment}

This work was supported by Fundamental Research Funds for the Central Universities (HUST: 2021GCRC058) and was part by the High Performance Computing (HPC) Platform of Huazhong University of Science and Technology where the computation is completed.

\printcredits

\bibliographystyle{cas-model2-names}

\bibliography{reference}

\end{document}